\documentclass[lettersize,journal]{IEEEtran}
\usepackage{amsmath,amsfonts}
\usepackage{algorithmic}
\usepackage{algorithm}
\usepackage{subcaption}
\usepackage{array}
\usepackage{textcomp}
\usepackage{stfloats}
\usepackage{url}
\usepackage{verbatim}
\usepackage{graphicx}
\usepackage{threeparttable}
\usepackage{hyperref}
\usepackage[numbers,sort&compress]{natbib}
\begin{document}

\title{Fair Uncertainty Quantification for Depression Prediction}

\author{Yonghong Li, Zheng Zhang and Xiuzhuang Zhou
\thanks{This work was funded in part by the National Natural Science Foundation of China under Grants 62576049 (\textit{Corresponding author: Xiuzhuang Zhou}).}
\thanks{Yonghong Li, Zheng Zhang and Xiuzhuang Zhou are with the School of Intelligent Engineering and Automation, Beijing University of Posts and Telecommunications, Beijing 100876, China  (Email: xiuzhuang.zhou@bupt.edu.cn).}
}

\markboth{Journal of \LaTeX\ Class Files,~Vol.~14, No.~8, August~2021}%
{Shell \MakeLowercase{\textit{et al.}}: A Sample Article Using IEEEtran.cls for IEEE Journals}


\maketitle

\begin{abstract}
Trustworthy depression prediction based on deep learning, incorporating both predictive reliability and algorithmic fairness across diverse demographic groups, is crucial for clinical applications. Recently, achieving reliable depression predictions through uncertainty quantification has attracted increasing attention. However, few studies have focused on the fairness of uncertainty quantification (UQ) in depression prediction. In this work, we investigate the algorithmic fairness of UQ, namely equal opportunity of coverage (EOC) fairness, and propose fair uncertainty quantification (FUQ) for depression prediction. FUQ pursues reliable and fair depression predictions through group-based analysis. Specifically, we first group all the participants by different sensitive attributes and leverage the group conditional conformal prediction method to quantify uncertainty within each demographic group, which provides a theoretically guaranteed and valid way to quantify uncertainty for depression prediction and facilitates the investigation of fairness across different demographic groups. Furthermore, we formulate the fairness of UQ as a constrained optimization problem under EOC constraints, and propose a fairness-aware optimization strategy to solve it. This enables the model to preserve predictive reliability while adapting to the heterogeneous uncertainty levels across demographic groups, thereby achieving optimal fairness. Through extensive evaluations on several visual and audio depression datasets, our approach demonstrates its effectiveness. 

\end{abstract}

\begin{IEEEkeywords}
Depression prediction, uncertainty quantification, algorithmic fairness, group conditional conformal prediction.
\end{IEEEkeywords}

\section{Introduction}
\IEEEPARstart{D}{epression} has emerged as the most prevalent mental disorder globally, serving as a leading cause of disability and a major risk factor for suicide-related mortality \cite{friedrich2017depression, world2017depression}. Beyond its psychological burden, depression leads to persistent low mood, cognitive decline, and social dysfunction \cite{piao2022alarming}. Moreover, it exhibits a bidirectional exacerbation mechanism with chronic physical illnesses \cite{lim2018prevalence, ismail2017prevalence}. 
Epidemiological estimates indicate that approximately 3.8\% of the global population suffers from depression, including 5\% of adults (4\% in males and 6\% in females), with a prevalence of 5.7\% among individuals aged 60 years and older. Notably, women are approximately 50\% more likely to experience depression than men, with over 10\% of pregnant and postpartum women suffering from depression. While the precise etiology of depression remains unclear, statistical analyses demonstrate that individuals who have experienced abuse, significant loss, or other major stressors are at an increased risk of developing the disorder. Additionally, women are consistently found to be at a greater risk of depression compared to men \cite{vos2015global, woody2017systematic, belmaker2008major, evans2018socio, institute2021global}.

\begin{figure}
    \centering
    \begin{subfigure}[b]{0.5\textwidth}
        \includegraphics[width=\textwidth]{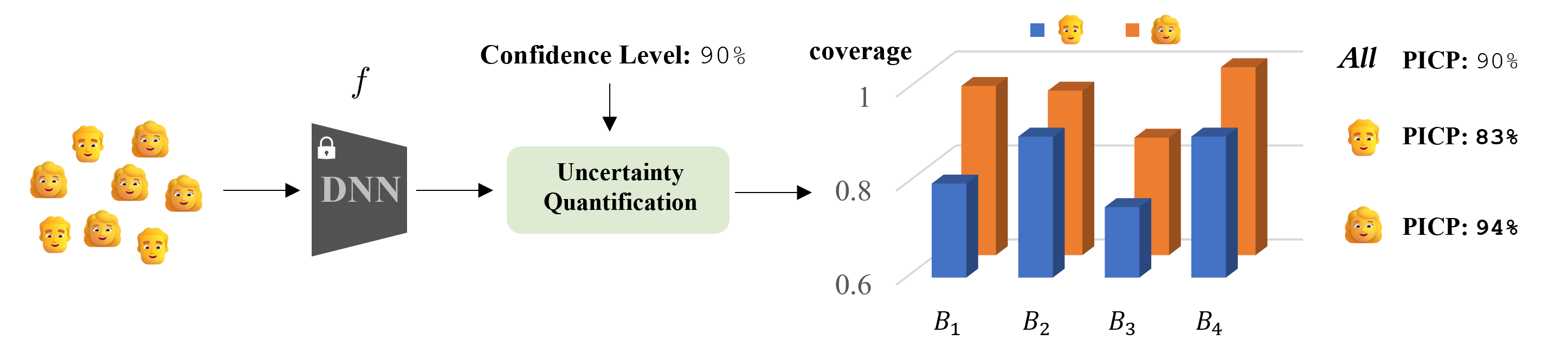}
        \caption{\footnotesize Depression prediction with UQ (e.g., CDP \cite{li2025conformal})}
        \label{uq}
    \end{subfigure}
    \hfill
    
    \begin{subfigure}[b]{0.5\textwidth}
        \includegraphics[width=\textwidth]{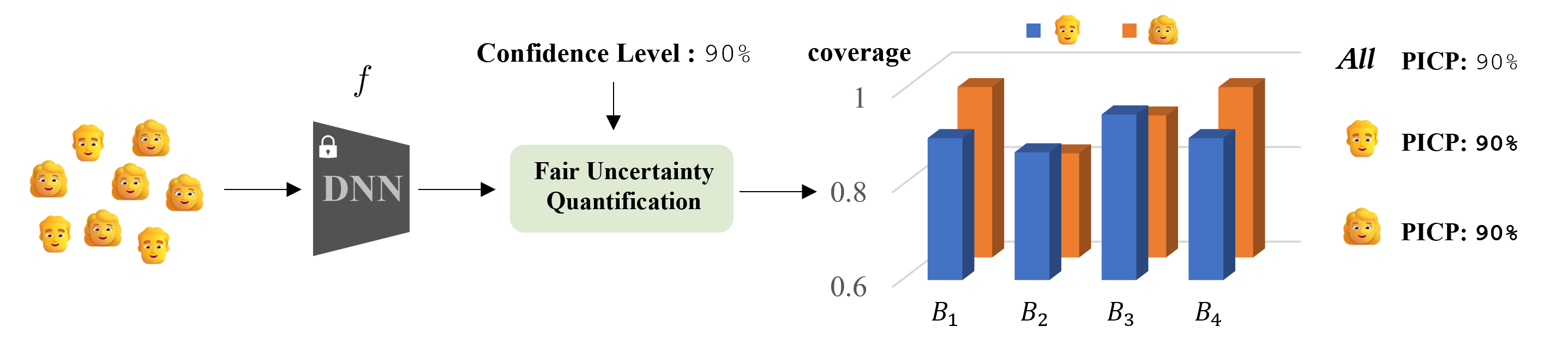}
        \caption{\footnotesize Depression prediction with FUQ}
        \label{fuq}
    \end{subfigure}
    \caption{Comparison of uncertainty quantification (UQ) approaches for depression prediction (using gender as the sensitive attribute): (a) Existing UQ provides statistically valid confidence intervals globally for \textbf{\textit{All}} participants, but exhibits disparate coverage rates across demographic groups for gender. (b) Fair uncertainty quantification (FUQ) achieves dual objectives: maintaining global statistical validity while ensuring equitable coverage across all subgroups. Here, $B$ indicates bins formed by evenly dividing participants according to depression level; this binning supports approximate conditional coverage in our interval construction.}
    \label{fig:fair}
\end{figure}

Traditional diagnostic frameworks for depression primarily rely on self-reported symptom assessments and clinical observations, yet they suffer from three major limitations \cite{beck1996beck,hamilton1986hamilton,kroenke2009phq}. First, symptom reporting is susceptible to patient stigma and cultural biases, leading to patients' reluctance to participate in diagnosis or an increased risk of misdiagnosis \cite{barney2006stigma}. Second, the diagnostic process is significantly delayed, requiring substantial time, effort, and resources from symptom onset to formal diagnosis, while the risk of relapse remains high even after treatment. Third, disparities in healthcare resource distribution result in the vast majority of patients in low- and middle-income countries lacking access to effective interventions \cite{ismail2017prevalence}.

In this context, deep learning-based Automatic Depression Recognition systems (ADRs) have emerged as a promising solution by behavioral data, including speech prosody, facial expression dynamics, and motor biomarkers, to enable a quantitative analysis of depressive traits \cite{williamson2013vocal,pan2023integrating,mao2022prediction,de2019depression,de2021mdn,de2020deep}. These systems capture depression-related features in naturalistic interaction settings, overcoming the spatial and temporal constraints of traditional diagnostic methods. Moreover, ADRs facilitate the development of dynamic monitoring frameworks, providing a transformative paradigm for early detection and timely intervention in depression. 

As research delves deeper, despite the outstanding capabilities for learning representations, researchers have found that the deep learning models are often untrustworthy \cite{guo2017calibration}. One major reason is that deep models are typically \textit{black boxes} and lack interpretability. Additionally, observations have shown that predictions often exhibit \textit{overconfidence}. In depression prediction, \cite{nouretdinov2011machine,li2025conformal} address this issue through uncertainty quantification (UQ). Calibrated predictions with UQ not only provide accurate predictions but also output predictions with the confidence level. 

Moreover, few studies have focused on the issue that data-driven deep learning algorithms are always prone to predictive bias, which varies across different groups and leads to unfair predictions. Such biases can come from various sources, including imbalanced data distribution, inherent algorithmic design, and task-specific requirements \cite{wang2022bias,mao2022biases,gunes2023fairness}. Kuzucu et al. \cite{kuzucu2024uncertainty} show that models produce systematically biased uncertainty estimates across demographic groups. Furthermore, empirical evidence shows that depression prediction is unfair across demographic groups, and such bias directly weakens the validity of uncertainty quantification, as shown in Fig. \ref{uq}. \cite{cheong2024fairrefuse,cheong2025u} seeks to achieve fair depression prediction by eliminating biases associated with specific attributes, but they overlook uncertainty quantification and calibration. Ensuring fairness in UQ is as critical as UQ itself; unfair uncertainty estimates can impose disproportionate and potentially catastrophic harms on a specific group. For example, if uncertainty is well calibrated for younger patients with depression but neglected for older patients, it would undermine trustworthiness for the latter and could lead to severe, irreversible harm.

However, achieving fair depression prediction within a UQ framework is nontrivial \cite{kuzucu2024uncertainty}. First, incorporating fairness considerations into existing UQ methods for depression recognition is neither intuitive nor straightforward. Most prevailing approaches prioritize aggregate statistical validity and, in doing so, neglect the demographic stratifications with sensitive attributes. Consequently, their uncertainty estimates tend to privilege majority groups while failing to adequately cover minority groups, thereby introducing unfairness. Second, stratifying the data by sensitive attributes substantially reduces the sample size within each group, which can, in turn, impair the statistical validity of the resulting UQ estimates. 
In this paper, we consider reliable depression prediction through effective UQ and fairness-aware optimization. The motivation for this work is summarized as follows: 
\begin{enumerate}
    \item Uncertainty quantification in depression prediction is inherently influenced by algorithmic design and uncertainty calibration. 
    \item From an algorithmic-fairness perspective, the valid quantification should be influenced only by model’s intrinsic predictive uncertainty rather than irrelevant attributes. For example, if the depression prediction’s confidence level is 85\% for males but 95\% for females, such a discrepancy is unfair.
\end{enumerate}

Given the absence of a well-established fairness definition in depression prediction, we focus on equal opportunity of coverage (EOC) fairness, an uncertainty-based fairness notion introduced in \cite{10.5555/3666122.3666462}. As illustrated in Fig. \ref{fig:fair}, compared to previous uncertainty quantification methods, EOC requires that the validity of UQ should not only satisfy the desired confidence rate across all participants but also maintain comparable validity across demographic groups (e.g., male and female). Our contributions can be summarized as follows:

1. We propose fair uncertainty quantification (FUQ), a framework for achieving fair and reliable depression predictions through group-based analysis. To the best of our knowledge, this work is the first to explore EOC-fairness in depression prediction through the lens of uncertainty quantification. 

2. In FUQ, we design a fairness-aware optimization strategy to preserve EOC fairness in depression prediction across demographic groups, while providing theoretically guaranteed and calibrated uncertainty estimates for each group.

3. Extensive experiments on AVEC 2013 \& 2014, DAIC-WOZ, and MODMA datasets demonstrate the effectiveness of our proposed method.

\begin{figure*}[t] 
    \centering 
    \includegraphics[width=\linewidth]{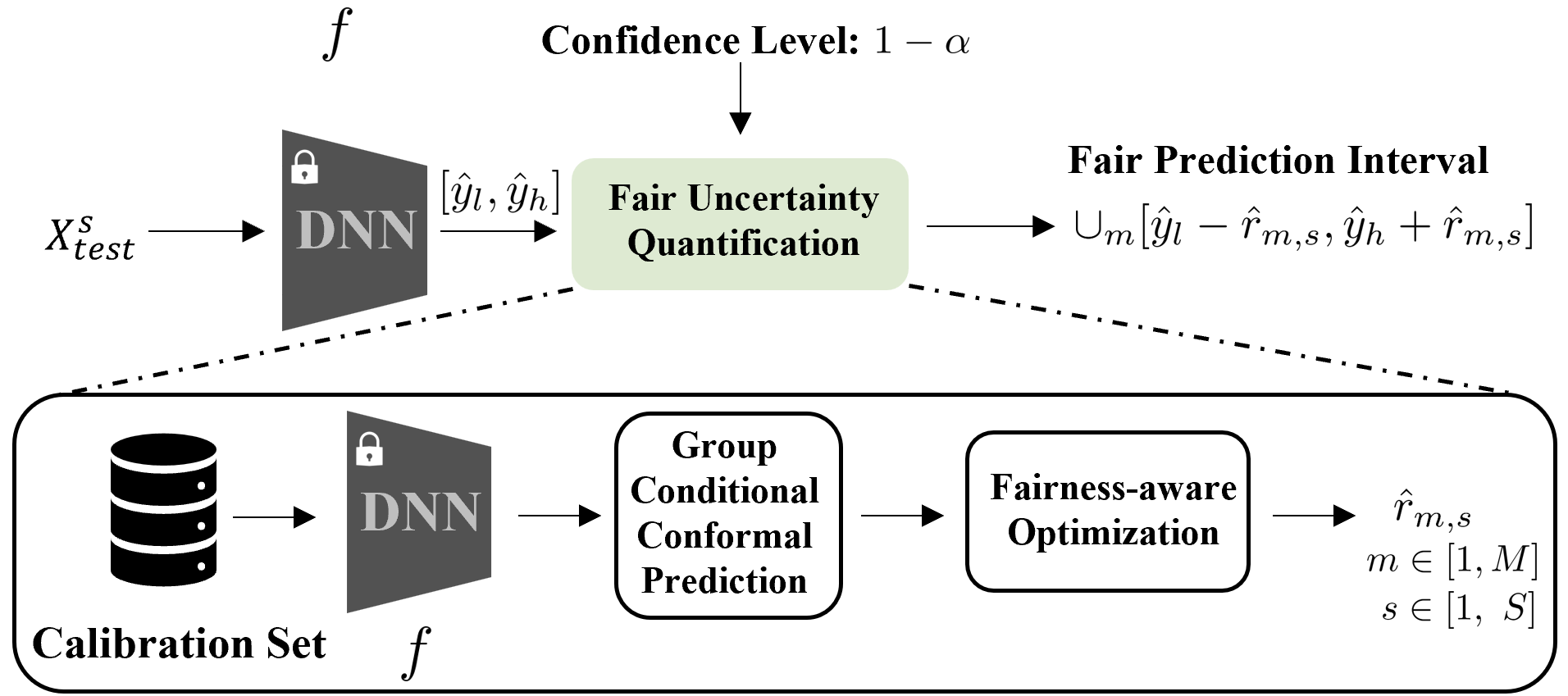} 
    \caption{The pipeline of FUQ for depression prediction. FUQ produces prediction interval that is both statistically valid and fair across demographic groups. It calibrates traditional prediction intervals (produced by quantile regression, for example) into fair intervals by coupling group conditional conformal prediction (ensuring reliable uncertainty estimates within group), with a fairness-aware optimization step (balancing inter-group coverage). $f$ denotes the quantile regression model, which outputs prediction intervals for uncertainty quantification (UQ), $M$ denotes the number of the dividing bins for conditional CP, and $S$ denotes the number of the sensitive attributes for grouping.} 
    \label{pipeline-FRDIP} 
\end{figure*}

\section{Related Work}
\subsection{Depression Recognition}
Depression recognition has attracted significant attention, initially originating from the AVEC challenge, which used the AVEC 2013 and 2014 facial videos to inspire a considerable community of researchers \cite{valstar2013avec,valstar2014avec}. Early efforts, such as those by Williamson et al., employed handcrafted audio features and trained a multivariate regression Gaussian mixture model (GMM), achieving notable success in the competition \cite{williamson2013vocal}. 

In recent years, deep learning has largely replaced traditional handcrafted features, leveraging its powerful representational capabilities to enhance recognition accuracy. Melo et al. advanced the field by proposing the multiscale spatiotemporal network (MSN) and maximization-differentiation network (MDN) to improve feature extraction for a more comprehensive depression representation \cite{de2020deep,de2021mdn}. Zhang et al. designed the lightweight multiscale temporal difference attention network (MTDAN) to address the high computational complexity of deep learning models, facilitating real-time clinical applications \cite{zhang2023mtdan}. Niu et al. estimated depression severity using facial keypoints representation sequences (FKRS) and action units representation sequences (AURS), avoiding the use of raw facial video data  \cite{niu2024depressionmlp}. Xu et al. proposed automatic facial depression recognition based on a two-stage framework, where short-term facial behaviors are first modeled, followed by video-level depression analysis using sequential graph representation (SEG) and spectral graph representation (SPG) \cite{xu2024two}. 

In addition to using a single modality, some studies have also focused on multimodal depression recognition. Yang et al. applied deep neural networks (DNNs) for multimodal fusion of audio, video, and text \cite{yang2017hybrid}. Mao et al. developed an attention-based multimodal representation by integrating Bi-LSTM with GloVe embeddings \cite{mao2022prediction}, while Uddin et al. introduced a video-speech multimodal prediction model that employs volume local directional structural pattern (VLDSP-based) dynamic feature descriptors for facial dynamics and temporal attention pooling (TAP) for segment-level audiovisual feature aggregation \cite{uddin2022deep}. Pan et al. proposed an audiovisual attention network (AVA-DepressNet), a multimodal framework that enhances audiovisual spatial-temporal features through attention mechanisms and optimizes the encoder-decoder structure using an adversarial multistage training strategy with facial structural priors \cite{pan2023integrating}. Fan et al. further extended multimodal depression detection by introducing transformer-based multimodal feature enhancement networks (TMFE) and employing a graph fusion network architecture to integrate features across different modalities \cite{fan2024transformer}. While advancements have reduced errors in depression recognition, most existing methods overlook predictive uncertainty and predictive bias, thereby undermining the trustworthiness of their predictions. Addressing these challenges remains crucial for developing trustworthy ADRs.

\subsection{Uncertainty and Bias in Affective Computing}

In the field of affective computing, UQ has also attracted significant attention. Zhou et al. utilized label distributions to model the uncertainty of predictions, where the smoother the learned label distribution, the greater the uncertainty \cite{9187982}. Lo et al. modeled uncertainty in low-resolution facial expressions through structured probabilistic embeddings \cite{lo2023modeling}. Ahn et al. enhanced representation learning by combining contrastive learning with features represented as probability distributions to model uncertainty \cite{ahn2024uncertainty}. Li et al. modeled uncertainty in depression prediction through interval predictions, where wider intervals indicate higher uncertainty \cite{li2025conformal}. These methods incorporate uncertainty analysis into affective computing through either heuristic approaches or uncertainty quantification. However, these methods prioritize the statistical performance of UQ and overlook predictive biases across demographic groups, resulting in inconsistent quantification performance. 

On the other hand, data-driven affective computing methods are often prone to data or algorithmic bias. Gratch et al. highlighted ethical considerations in affective computing, emphasizing the biases arising from insufficient diversity in dataset attributes \cite{verhoef2023towards,gratch2024guest}. Similar observations are evident in related studies. Li et al. investigated biases in facial expression recognition and proposed the emotion-conditional adaptation network (ECAN) to learn domain-invariant facial representations \cite{li2020deeper}. Adarsh et al. combined support vector machines (SVM) and K-Nearest Neighbors (KNN) in an ensemble model to address age-related prediction biases, striving for fair depression detection from social media data \cite{adarsh2023fair}. Additionally, Dang et al. addressed prediction unfair biases in depression prediction using machine learning, proposing population sensitivity-guided threshold adjustment (PSTA) to achieve fair prediction \cite{dang2024fairness}. Ensuring fairness is essential for developing ADRs that are not only accurate but also trustworthy and ethically responsible. Notably, Cheong et al. introduced the FAIRREFUSE framework, a dynamic referee-guided causal approach employing causal intervention and individual fairness-guided fusion to mitigate gender biases in multimodal depression recognition \cite{cheong2024fairrefuse}. Additionally, they proposed a gender-based task-reweighting approach leveraging uncertainty estimation to enhance performance and fairness in depression detection \cite{cheong2025u}. 

Prior studies have addressed predictive bias in depression recognition and have maintained fairness by mitigating such bias. Nevertheless, incorporating fairness into UQ remains challenging. Traditional UQ performance metrics emphasize overall statistical validity, which depends on aggregate sample size and therefore tend to overlook disparities associated with sensitive attributes; furthermore, stratifying data by these attributes reduces group sample sizes and consequently degrades UQ validity. Cheong et al. share a partially similar underlying motivation with ours \cite{cheong2025u}, concentrating on fairness-aware depression detection via uncertainty-driven heuristics formulated as a classification task. By contrast, our study integrates fairness into rigorously calibrated UQ, and advances reliable depression prediction formulated as a regression task.


\section{Method}


\subsection{UQ for Depression Prediction}
Let the depression dataset be divided into a training set $\mathcal{D}_{train}$, a calibration set $\mathcal{D}_{cal}$, and a test set $\mathcal{D}_{test}$. We denote $s$ as the sensitive attribute  (such as age, gender) of the input $X$, and $y$ as the depression score. Note that $S$ denotes the number of categories for a given attribute (e.g., for gender, $S=2:$ male and female), not the number of distinct attributes. For the traditional depression prediction, we train a model by minimizing the loss (such as MAE, MSE) as the learning objective on the $\mathcal{D}_{train}$. However, the model does not incorporate uncertainty quantification capabilities. This work trains a quantile regression model to perform preliminary uncertainty quantification. Specifically, during training, we apply the pinball loss as the learning objective to achieve quantile prediction. 

\begin{equation}\label{pinball}
\mathcal{L}_{pinball} =
\begin{cases}
q \cdot (y - \hat{y}_q), & \text{if } y \geq \hat{y}_q \\
(1 - q) \cdot (\hat{y}_q - y), & \text{if } y < \hat{y}_q
\end{cases}
\end{equation}
where $ q \in [0, 1]$, $\hat{y}_q$ denotes the $q$-th quantile of the prediction. This enables the model to learn the lower and upper quantile bounds $\hat{y}_l$ and $\hat{y}_h$, allowing for the construction of prediction intervals that represent uncertainty. To further calibrate the performance of uncertainty quantification, and following \cite{li2025conformal}, we adopt conformalized quantile regression (CQR) \cite{romano2019conformalized} for uncertainty quantification in depression prediction.  
 Let $\alpha$ denote the miscoverage rate. On the $\mathcal{D}_{cal}$, we then utilize the ($1-\alpha$)-th \{$\max \{\hat{y}_{\alpha/2} - y, y - \hat{y}_{1 - \alpha/2}\}$\} to obtain $\beta$, the $1-\alpha$ quantile of the conformal scores. The valid $1-\alpha$ confidence interval is denoted as $C(X) = [\hat{y}_{\alpha/2} - \beta, \hat{y}_{1 - \alpha/2} + \beta]$ for $X \in \mathcal{D}_{test}$. CQR yields a ($1-\alpha$) level confidence interval $C$ with a marginal coverage guarantee, under exchangeability, 
\begin{equation}\label{marginal}
P(y \in C(X)) \geq 1 - \alpha
\end{equation}

While valid, marginal coverage can be further refined to better approximate conditional coverage (even though \cite{vovk2012conditional} shows that conditional coverage is hard to achieve, it can be approximated). 
\begin{equation}\label{marginal}
P(y \in C(X)|X = X_i) \geq 1 - \alpha
\end{equation}

The current limitation is that CQR uses the fixed global calibration factor $\beta$ that does not vary with a test sample’s depression severity. To approximate conditional coverage, we adopt a simple scheme by partitioning participants into $M$ equal-mass bins by depression severity, $B_m = [l_m,u_m)$, $m \in [1,M]$, and $l, u$ denote the lower and upper bounds of the bin respectively, and computing a group-specific calibration factor $\beta_m$ for each group. Thus, the prediction intervals can be regarded as achieving approximate conditional coverage across different levels of depression severity. However, empirical observations reveal that coverage varies across sensitive attribute groups, indicating non-uniform effectiveness. Some groups demonstrate over-coverage, while others do not meet the required coverage guarantee. Therefore, we pursue equal opportunity of coverage (EOC).

\begin{algorithm}[t]
\caption{FUQ for Depression Prediction}\label{alg:alg1}
\begin{algorithmic}[1]
\REQUIRE $\mathcal{D}_{cal} = \{X_i^{s},y_i\}_{i = 1}^N, s \in [1,S]$: calibration set;\newline
$X_{test}^s$: test data;\newline
$f$: well-trained quantile model on the training set;\\
$\mathcal{Q}(\cdot,\cdot)$: quantile estimator;\newline
$\alpha \in (0,1)$: miscoverage rate;\newline
$B_1,B_2,...,B_M$: $M$ equal-mass bins; \\
$l_m, u_m$: lower and upper bounds of $B_m$; \\
$X_{test}^s$: test data with a specific attribute $s$. 
\ENSURE Fair and reliable prediction $C(X_{test}^s)$.\\      

\texttt{// Calculate the $1-\alpha$ intervals of CQR and initialize subgroup coverage $\beta_{m,s}$ }
\FOR{$i = 1$ \textbf{to} $N$}
    \STATE $r_i = \max\{f_{\alpha/2}(X_i^{s}) - y_i, y_i - f_{1-\alpha/2}(X_i^{s})\}$
\ENDFOR
\STATE $\hat{r} = \mathcal{Q}(1-\alpha,r)$ \\
\FOR{$i = 1$ \textbf{to} $N$}
    \STATE $C(X_i^{s}) = [f_{\alpha/2}(X_i^{s}) - \hat{r}, f_{1-\alpha/2}(X_i^{s}) + \hat{r}]$ \\
    \FOR{$m = 1$ \textbf{to} $M$}
        \FOR{$s' = 1$ \textbf{to} $S$}
            \STATE $\beta_{m,s} \gets \frac{\mathbf{1}\{y_i \in (B_m \cap C(X_i^{s}))\} * \mathbf{1}\{s = s'\}}{|B_m|}$
        \ENDFOR
    \ENDFOR\\
\ENDFOR

\STATE Using Eqs. \ref{interval}-\ref{rs1} to obtain $\hat{r}_{m,s}$

\texttt{// Calculate the confidence interval}
\FOR{$m = 1$ \textbf{to} $M$}
    \STATE $C_{m}(X_{test}^s) = [f_{\alpha/2}(X_{test}^s) - \hat{r}_{m,s}, f_{1-\alpha/2}(X_{test}^s) + \hat{r}_{m,s}] \cap [l_m, u_m)$ \\
    \STATE $C(X_{test}^s) = \bigcup_{m} C_{m}(X_{test}^s)$
\ENDFOR
\RETURN $C(X_{test}^s)$ \\

\end{algorithmic}
\end{algorithm}

\subsection{FUQ for Depression Prediction}
Achieving EOC entails two conditions. First, the coverage across different sensitive attribute groups should be equal (we illustrate with groups $s_1$ and $s_2$, though practical applications may involve more). 
\begin{equation}
\label{eoc}
P(y \in C(X^s)|s = s_1) = P(y \in C(X^s)|s = s_2) \geq 1 - \alpha
\end{equation}

Second, each group should meet the expected coverage. 
The most straightforward method is to group the data based on sensitive attributes and apply CQR to output confidence intervals separately within each group. While this sounds reasonable, grouping significantly sharpens the available data for each group, thereby diminishing the statistical validity guarantees of UQ.

We leverage a group conditional conformal prediction with equal-mass binning to maximize within-group validity and to facilitate the implementation of fairness interventions \cite{10.5555/3666122.3666462}. 
Specifically, it divides all participants in the calibration set into $M$ equal-mass bins, $B_m = [l_m,u_m)$, $m \in [1,M]$, and $l, u$ denote the lower and upper bounds of the bin respectively, with respect to the label $y$. This bin-dividing method ensures that each bin is given equal importance while maximizing the size of each bin, and it is beneficial to permit the independent optimization of coverage for $M$ bins, thereby ensuring marginal coverage while maintaining the interval widths at their narrowest possible point. Since each bin in group conditional conformal prediction is optimized independently, we still need a fairness-aware optimization for the prediction interval while keeping the interval widths as narrow as possible to be informative. That is 
\begin{equation}
\begin{aligned}
\min & \sum_{m, i \in [1,\mathcal{D}_{cal}]} \frac{|C_m(X_i)|}{|\mathcal{D}_{cal}|} \\
\text{s.t.} \quad & 
\begin{cases}
\sum_m \frac{\beta_{m,s}}{M} = 1 - \alpha, \quad s \in [1,S],  \\
\beta_{m,s} \in [0, 1], \quad m \in [1,M].
\end{cases}
\end{aligned}
\end{equation}

By leveraging the gradient of coverage with respect to conformal scores, we optimize the confidence intervals across different attributes, resulting in fairer UQ for depression prediction, as shown in Algorithm \ref{alg:alg1}. 

Two key considerations are required to achieve this optimization objective:
\begin{enumerate}
    \item \textbf{Maintain overall validity:} Fairness interventions should not compromise predictive validity; the method of UQ should maintain its overall validity while simultaneously promoting fairness. 
    \item \textbf{Equal coverage across sensitive attributes:} The coverage rates should be consistent across different sensitive attribute groups, ensuring a uniform coverage rate of $1 - \alpha$.
\end{enumerate}

In order to achieve this optimal fairness-aware UQ, it is necessary to transform the optimizing sub-coverage $\beta_{m,s}$ into optimizing the quantile sub-threshold $\hat{r}_{m,s}$ of conformal scores within the group conditional conformal prediction framework. To guarantee overall statistical validity, we initially disregarded sensitive attributes and acquired the quantile $\hat{r}$ of conformal scores $\beta_{m}$ for the complete calibration set. For each sensitive-attribute bin, the initial sub-coverage $\beta_{m,s}^0$ is computed using $\hat{r}_{m,s} = \hat{r}$. $\beta_{m,s}^0$ serves as the starting point for the fairness-aware optimization. 
$\hat{r}_{m,s_2}$ (ideal coverage is $1-\alpha$). 
\begin{equation}\label{interval}
C(X_i^s) = \left[ f_{\alpha/2}(X_i^s) - \hat{r}_{m,s}, 
                f_{1-\alpha/2}(X_i^s) + \hat{r}_{m,s} \right]
\end{equation}
\begin{equation}\label{beta}
\beta_{m,s} \gets \frac{\mathbf{1}\{y_i \in (B_m \cap C(X_i^{s}))\} }{|B_m|}
\end{equation}
Although the initial sub-coverage may not be perfectly fair across demographic groups, it still satisfies overall statistical validity, which is the primary advantage of this initialization. If we straightly perform group calibration, with a significantly reduced sample size, the validity guarantee of the CP is also reduced. 

After sub-coverage initialization, we need to compare the coverage of each sensitive attribute, i.e., compute the average coverage of $M$ bins. For over-covered groups of sensitive attributes, for example $s_1$, we need to decrease $\hat{r}_{m,s_1}$. Conversely, under-covered groups of sensitive attributes, for example $s_2$, need to increase $\hat{r}_{m,s_2}$. 
As for the updating method, we need to use the slope of $\beta_{m,s}$, instead of the gradient, since we cannot guarantee the differentiable property. For $m = 1,...,M$, we can acquire $\nabla \hat{r}_m^{-}$ and $\nabla \hat{r}_m^{+}$. 
\begin{equation}\label{r-}
\nabla \hat{r}_m^{-} \gets \frac{\mathcal{Q}(\beta_{m,s_{0}}, r_{m,s_{0}}) - 
                \mathcal{Q}(\beta_{m,s_{0}}^{-}, r_{m,s_{0}})}{|B_{m,s_{0}}|} 
\end{equation}

\begin{equation}\label{r+}
\nabla \hat{r}_m^{+} \gets \frac{\mathcal{Q}(\beta_{m,s_{1}}^{+}, r_{m,s_{1}}) - 
                \mathcal{Q}(\beta_{m,s_{1}}^{e}, r_{m,s_{1}})}{|B_{m,s_{1}}|} 
\end{equation}

Furthermore, we can think of the overall validity guarantee as covering $(1-\alpha) * |\mathcal{D}_{cal}|$ samples on the calibration set. The over-coverage group should cover one fewer sample, while the under-coverage group should cover one more sample. In order to obtain optimal interval widths, it is necessary to ensure that the over-coverage group maximizes the slope, $\nabla \hat{r}^{-} = \max \nabla \hat{r}_m^{-}$, when it covers one fewer sample (i.e., decreases the interval width as much as possible). The under-coverage group minimizes the slope, $\nabla \hat{r}^{+} = \min \nabla \hat{r}_m^{+}$, when it covers one more sample (i.e., increases the interval width as little as possible). We can represent the effect of covering one less (or one more) sample on the width of the interval by computing the slope of $\beta_{m,s}$ and acquiring $\hat{r}_{m,s}$.
\begin{equation}\label{rs0}
\hat{r}_{m,s_1} = \hat{r}_{m,s_1} - \nabla \hat{r}^{-}
\end{equation}
\begin{equation}\label{rs1}
\hat{r}_{m,s_2} = \hat{r}_{m,s_2} + \nabla \hat{r}^{+}
\end{equation}

We update $\hat{r}_{m,s}$ using the fixed-point iterations in Eqs. \ref{interval}-\ref{rs1} before achieving EOC.The natural termination condition for optimization is that both the under-coverage and over-coverage groups achieve the desired coverage, or that the slope of the under-coverage group is greater than that of the over-coverage group.

During evaluation, for the participant $X_{test}^s$ with a given attribute $s$, we hypothetically assign the depression score to each severity bin $B_m$; for each $B_m$, we construct the FUQ interval with the corresponding calibration threshold $\hat{r}_{m,s}$ and enforce feasibility by intersecting with the bin bounds. The overall prediction interval is obtained as their union.
 
\begin{table*}
\centering
\caption{Prediction results on different attribute groups}
\label{baseline}{
\begin{tabular}{c c c c c c c c}   
\hline
Dataset & Group & Modality & Val MAE & Val RMSE & Test MAE & Test RMSE  \\
\hline
AVEC 2013 & All & Video & 6.44 & 7.95  &6.52 & 8.10 \\ 
AVEC 2013 & Female & Video &6.76&8.23&6.55&8.57  \\ 
AVEC 2013 & Male & Video &6.17&7.65&6.03&7.86  \\ 
\hline
AVEC 2014 & All & Video & 5.53 & 7.71  &5.86 & 7.49 \\ 
AVEC 2014 & Female & Video & 5.12 & 6.40&5.12&6.92 \\
AVEC 2014 & Male &  Video & 6.68 & 8.33& 7.15&8.39 \\
\hline
DAIC-WOZ & All & Audio & 5.24 & 6.04  & 5.19 & 6.08  \\ 
DAIC-WOZ & Female & Audio & 4.75 & 6.05 & 5.48 & 6.56 \\ 
DAIC-WOZ & Male & Audio &5.54  & 6.35 &5.43  &6.23  \\ 
\hline
MODMA & All & Audio & 5.62 & 7.15 & 5.64 & 6.95 \\ 
MODMA & Female & Audio & 8.34 & 9.40 & 7.84 & 8.83 \\ 
MODMA & Male & Audio &5.81 & 7.62 & 5.87 & 7.51 \\ 
MODMA & Young & Audio & 5.42 & 6.88 & 5.50 & 6.72 \\ 
MODMA & Old & Audio & 6.35 & 7.68 & 6.03 & 7.69 \\ 

\hline
\end{tabular}} 
\end{table*}

\section{EXPERIMENTS}
To validate the effectiveness of FUQ, we conducted experiments on several common datasets, including AVEC \cite{valstar2013avec,valstar2014avec}, DAIC-WOZ \cite{gratch2014distress}, and MODMA \cite{cai2022multi}. For generality, we covered audio and video modalities and relied on methods with public code and pretrained weights (or otherwise easy to reproduce). We then compared FUQ with established trustworthy prediction baselines and further analyzed the fairness in UQ.

\subsection{Dataset}
AVEC 2013 \& 2014 \cite{valstar2013avec,valstar2014avec}: These datasets are subsets of the AViD-Corpus. The main difference is that AVEC 2013 includes 150 videos, while AVEC 2014, building on 2013, is divided into two tasks: Northwind and Freeform, each containing 150 videos. In AVEC 2013, the training set consists of 50 videos, the validation set contains 50 videos, and 50 videos are used for testing. For AVEC 2014, 200 videos (a combination of the original training and validation sets, male: female = 36: 64) were used for training, with 50 videos (the original Northwind test set, male: female = 18: 32) as the validation set, and 50 videos (the original Freeform test set, male: female = 18: 32) as the test set. We divided the original videos into consecutive video clips, each consisting of 16 frames. For AVEC 2013, there is no overlap between consecutive clips, while for AVEC 2014, there is an 8-frame overlap. Additionally, we used OpenFace \cite{baltruvsaitis2016openface} for inter-frame facial alignment, aligning faces based on facial landmarks. 

DAIC-WOZ \cite{gratch2014distress}: As part of the AVEC 2017 Challenge dataset, this dataset includes video and audio interview records of participants interacting with the virtual agent Ellie, with PHQ-8 questionnaire scores serving as labels. It consists of 189 participants, with 107 participants designated as the training set (male: female = 63: 44), 35 participants as the validation set (male: female = 16: 19), and 47 participants as the test set. In our experiments, we focus solely on the audio modality, extracting participant audio over time frames and utilizing log-Mel spectrogram features. The settings are as follows: Hanning window is set to 1024, audio sample rate is 16 kHz, hop length is 128, and Mel filters are 64.

MODMA \cite{cai2022multi}: This is a multimodal depression recognition dataset proposed by Lanzhou University, containing both audio and EEG modalities. The dataset consists of 52 participants (23 major depressive disorder subjects and 29 healthy control subjects), with ages ranging from 18 to 25 years ($<$30: $\geq$30 = 1: 1), and PHQ-9 questionnaire scores as labels. In this experiment, we focus on the audio modality and randomly split the data into training, validation, and test sets using a 3: 1: 1 ratio. We extracted participant audio over time frames and utilized log-Mel spectrogram features. The settings are as follows: Hanning window is set to 1024, audio sample rate is 44.1 kHz, hop length is 128, and Mel filters are 64.

\subsection{Experimental Setup}
\subsubsection{Predictive Model}

For the predictive model on AVEC 13 \& 14, we adopted a C3D network. Following previous studies \cite{li2025memrank}, we adjusted the penultimate fully connected (FC) layer to 64. For the DAIC-WOZ audio dataset, we designed a 1D CNN architecture inspired by \cite{lin2020towards}. We utilized the Adam optimizer with a weight decay of 0.0001 and a learning rate of 0.0001. Similarly, for the MODMA audio dataset, we adopted a network configuration similar to that of the DAIC-WOZ dataset.

For all three networks, the final FC layer consists of 99 units, corresponding to the quantile settings from 0.01 to 0.99 for pinball loss. This design enables the training of a multi-output quantile regression model. We consider the output corresponding to the 50\% quantile as the model’s prediction. Alternatively, multiple single-output quantile regression models could also be trained separately. With well training, we can obtain the quantile regression model $f$.

\subsubsection{Trustworthy Depression Prediction}
In regression tasks, CP \cite{vovk1999machine} and CQR \cite{romano2019conformalized} can obtain valid prediction intervals. CP does not require retraining of the model and can be utilized for vanilla regression. The disadvantage of CP is that the width of the prediction interval is fixed, which is not suitable for individuals with varying features, such as depressed patients. In contrast, CQR necessitates retraining the original single‑output depression prediction model, converting it into a multi‑output quantile regression model, thereby yielding adaptive prediction distributions for different individuals. This makes the prediction intervals align more closely with the true uncertainty of depression predictions. However, both methods ignore fairness. Although group-wise partitioning can be used to further mitigate disparities across groups, it often results in a reduced sample size within each group, which can compromise the statistical effectiveness of post-hoc methods such as conformal prediction. 
CFQR \cite{liu2022conformalized} considers the fairness of the quantile estimates in CQR under the requirement of Demographic Parity. However, applying fairness post-processing using the Wasserstein barycenter may introduce excessive risk \cite{chzhen2022minimax}. In contrast, BFQR \cite{10.5555/3666122.3666462} leverages a valid interval optimization strategy to adjust the interval width, achieving improved calibration performance. However, the optimization strategy in BFQR primarily focuses more on overall effectiveness. In this context, our FUQ further refines the optimization strategy by incorporating a fairness-aware optimization framework for prediction intervals, thereby achieving a more equitable coverage rate across demographic groups.

\begin{table*}
\centering
\caption{Predictive reliability on gender group for different methods}
\label{fair}{
\begin{tabular}{c c c c c c c }   
\hline
Method & Dataset & PICP(\%) & MPIW & PICP(s = 1) & PICP (s = 2)& PICP Gap $\downarrow$ \\
\hline
CP \cite{vovk1999machine}  &  AVEC 2013 & 91.25 &  22.81 & 87.22 & 90.00 & 2.78   \\ 
CQR \cite{romano2019conformalized}  &  AVEC 2013& 92.37 & 26.70 & 90.73 & 89.75 & 0.98\\
CFQR \cite{liu2022conformalized} &  AVEC 2013 & 93.43 & 27.86 & 91.42 & 93.46 & 2.04 \\
BFQR\cite{10.5555/3666122.3666462}  &  AVEC 2013 &92.13& 24.48 & 91.45 & 93.24 & 1.79  \\ 
Ours  &  AVEC 2013 & 90.14 & 22.68 & 90.40& 90.02 & \textbf{0.38} \\
\hline
CP \cite{vovk1999machine}  &  AVEC 2014 & 92.91 &  23.07 & 88.31 & 86.74 & 1.57   \\ 
CQR \cite{romano2019conformalized}  &  AVEC 2014& 91.67 & 22.41 & 88.31 & 87.16 & 1.15\\
CFQR \cite{liu2022conformalized} &  AVEC 2014 & 93.28 & 23.69 & 87.16 & 89.02 & 1.86 \\
BFQR \cite{10.5555/3666122.3666462}  &  AVEC 2014 &90.43& 19.15 & 90.73 & 89.73 & 1.00  \\ 
Ours  &  AVEC 2014 & 90.35 & 19.77 & 90.21& 90.69 & \textbf{0.48} \\
\hline
CP \cite{vovk1999machine}  & DAIC-WOZ & 93.36 &  20.02 & 91.33 & 95.57 & 4.24   \\ 
CQR \cite{romano2019conformalized}  & DAIC-WOZ & 94.83 & 19.52 & 93.08 & 95.31 & 2.23\\
CFQR \cite{liu2022conformalized} & DAIC-WOZ & 95.23 & 20.05 & 96.02 & 94.16 & 1.86 \\
BFQR \cite{10.5555/3666122.3666462}  & DAIC-WOZ & 95.05 & 22.01 & 94.23 & 97.00 & 2.77  \\ 
Ours  & DAIC-WOZ & 92.43 & 19.77 & 92.21& 92.69 & \textbf{0.48} \\
\hline
CP \cite{vovk1999machine}  & MODMA  & 90.16 & 19.73  & 90.59 & 91.35 & 0.76   \\ 
CQR \cite{romano2019conformalized}  & MODMA & 89.73 & 19.63 & 90.72 & 90.39 & 0.33\\
CFQR \cite{liu2022conformalized} & MODMA  & 93.26 & 20.25 & 94.22 & 92.39 & 1.83 \\
BFQR \cite{10.5555/3666122.3666462}  & MODMA  & 98.22 & 21.19 & 95.44 & 96.04 & 0.6 \\ 
Ours  & MODMA  & 90.94 & 19.92 & 90.79& 91.01 & \textbf{0.22} \\
\hline

$\alpha = 0.1, s = 1$ for female, $s = 2$ for male.
\end{tabular}}
\end{table*}

\begin{table*}
\centering
\caption{Predictive reliability on age group for different methods}
\label{fair-age}{
\begin{tabular}{c c c c c c c }   
\hline
Method & Dataset & PICP(\%) & MPIW & PICP(s = 1) & PICP (s = 2)& PICP Gap $\downarrow$\\
\hline
CP \cite{vovk1999machine}  & MODMA  & 92.59 & 20.51  & 93.62 & 91.44 & 2.18   \\ 
CQR \cite{romano2019conformalized}  & MODMA & 93.26 & 19.99 & 94.20 & 91.73 & 2.47\\
CFQR \cite{liu2022conformalized} & MODMA  & 95.67 & 21.32 & 96.34 & 94.37 & 1.97 \\
BFQR \cite{10.5555/3666122.3666462}  & MODMA  & 94.13 & 20.27 & 95.13 & 93.21 & 1.92 \\ 
Ours  & MODMA  & 91.05 & 18.55 & 90.82& 91.54 & \textbf{0.72} \\
\hline

$\alpha = 0.1, s = 1$ for young, $s = 2$ for old.
\end{tabular}}
\end{table*}

\subsubsection{Metrics}
FUQ requires performance evaluation from three perspectives. For accuracy, we use Mean Absolute Error (MAE) and Root Mean Squared Error (RMSE) as evaluation metrics. For the confidence of UQ, the quality of the predicted intervals is assessed using Prediction Interval Coverage Probability (PICP) and Mean Prediction Interval Width (MPIW).

\begin{equation}\label{picp}
PICP = \frac{1}{|\mathcal{D}_{test}|}\sum_{i = 1}^{|\mathcal{D}_{test}|}\mathbf{1}(y_i\in\mathcal{C}_{\alpha}(X_i))
\end{equation} 
\begin{equation}\label{mpiw}
MPIW =  \frac{1}{|\mathcal{D}_{test}|}\sum_{i = 1}^{|\mathcal{D}_{test}|}|\hat{y}_{i,h} - \hat{y}_{i,l}|
\end{equation}
where $X_i \in \mathcal{D}_{test}$ and $\mathcal{C}_{\alpha}(X_i) = [\hat{y}_{i,l}, \hat{y}_{i,h}]$.
\begin{figure*}                
  \centering
  \begin{subfigure}[b]{0.45\textwidth}
    \centering
    \includegraphics[width=\linewidth]{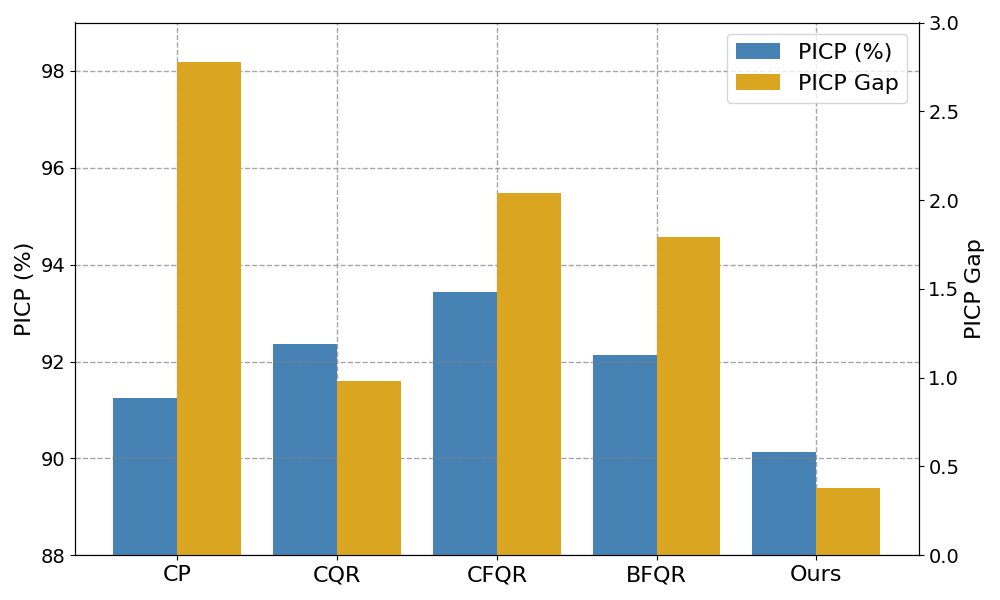}
    \caption{\footnotesize AVEC 2013}
    \label{fig:triple-A}
  \end{subfigure}
  \begin{subfigure}[b]{0.45\textwidth}
    \centering
    \includegraphics[width=\linewidth]{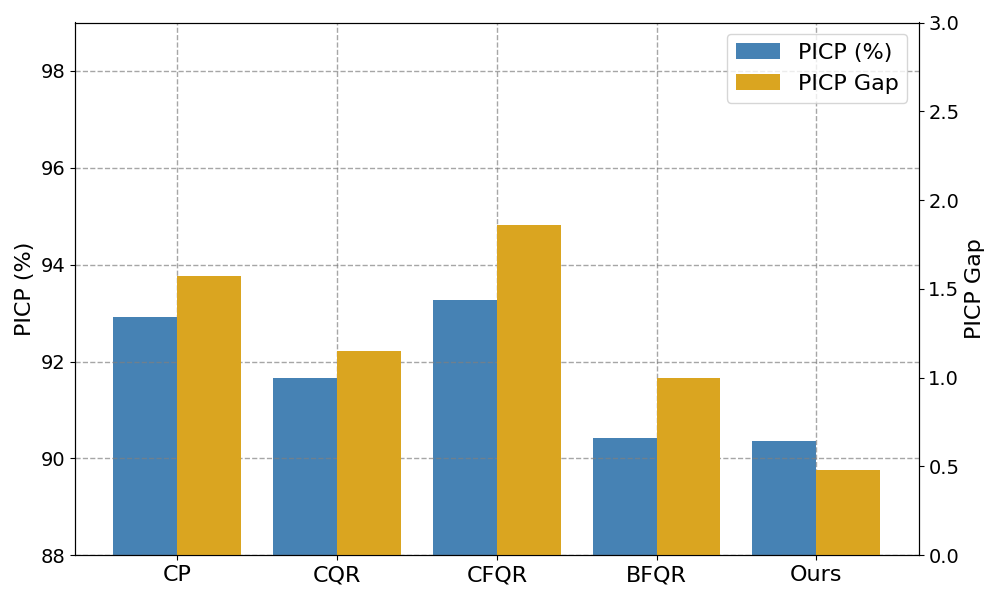}
    \caption{\footnotesize AVEC 2014}
    \label{fig:triple-B}
  \end{subfigure}
  \begin{subfigure}[b]{0.45\textwidth}
    \centering
    \includegraphics[width=\linewidth]{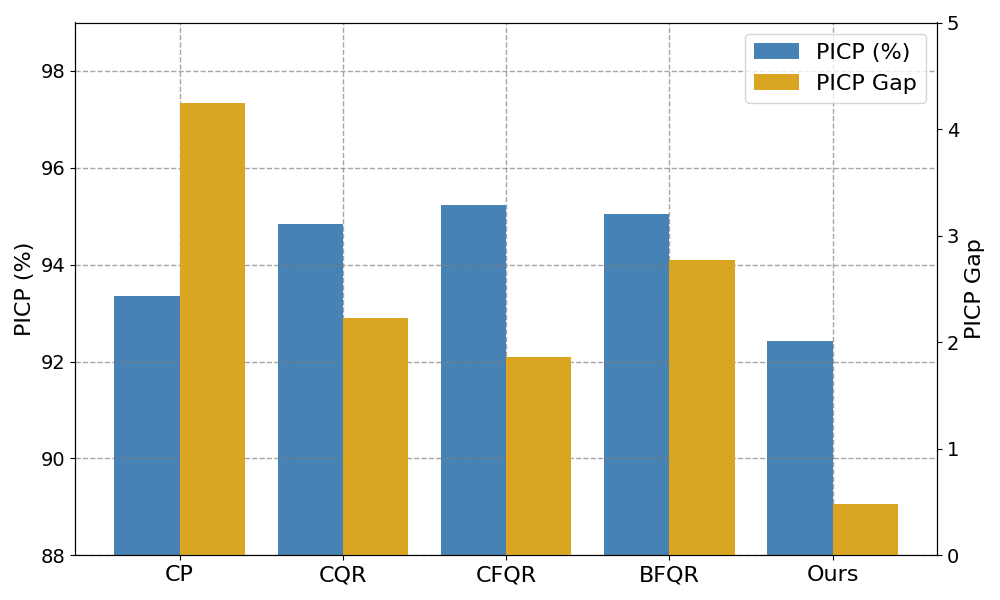}
    \caption{\footnotesize DAIC-WOZ}
    \label{fig:triple-C}
    \end{subfigure}
  \begin{subfigure}[b]{0.45\textwidth}
    \centering
    \includegraphics[width=\linewidth]{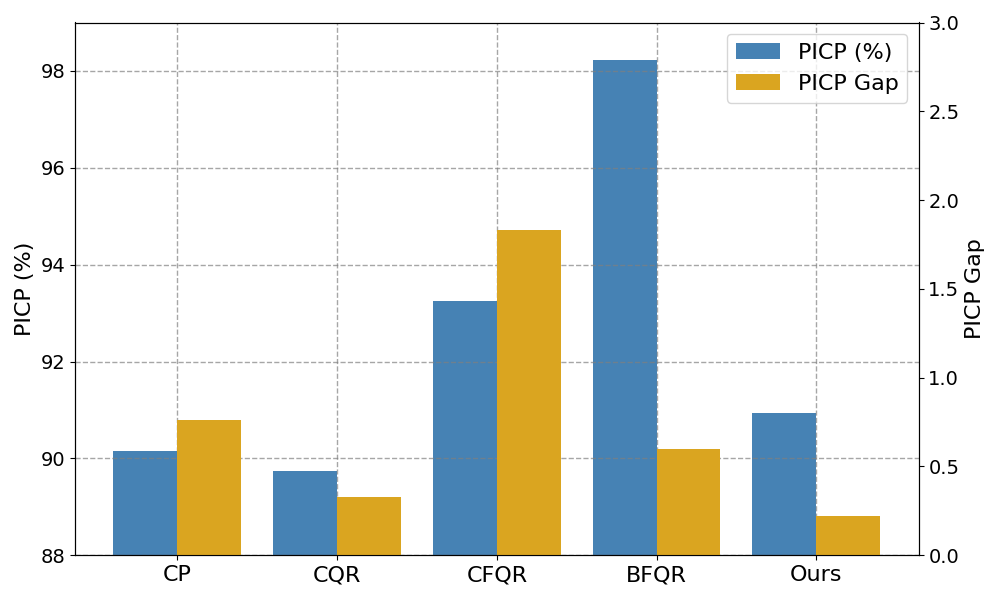}
    \caption{\footnotesize MODMA}
    \label{fig:triple-C}
    \end{subfigure}

  \caption{Performance of different methods on gender group across various datasets. At the 90 \% confidence level ($\alpha = 0.1$), our method yields the smallest group-wise coverage gap.}
  \label{fig:triple}
\end{figure*}

\begin{figure}
  \centering
  \begin{subfigure}[b]{0.45\textwidth}
    \includegraphics[width=\textwidth]{modma_gender.png}
    \caption{\footnotesize Gender Group}
    \label{fig:left}
  \end{subfigure}
  \hfill
  \begin{subfigure}[b]{0.45\textwidth}
    \includegraphics[width=\textwidth]{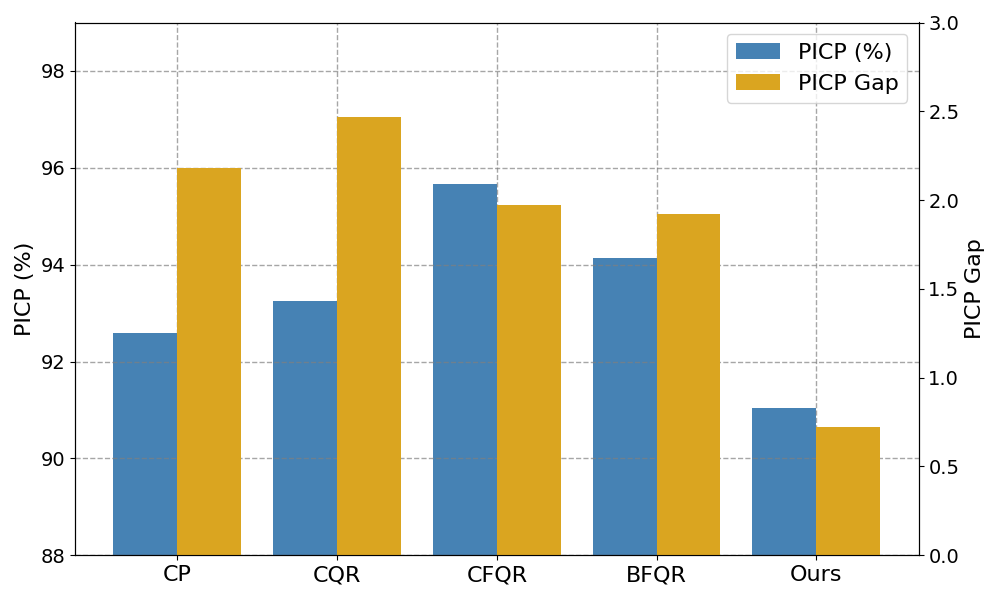}
    \caption{\footnotesize Age Group}
    \label{fig:right}
  \end{subfigure}
  \caption{Performance of various methods on the MOMDA test set across different demographic groups: (a) gender and (b) age. At the 90 \% confidence level ($\alpha = 0.1$), our method yields the smallest group-wise coverage gap.}
  \label{fig:twosubfigs}
\end{figure}

For the EOC fairness in predictions, a fair prediction interval should ensure that the PICP across different groups remains close to the expected $1-\alpha$ coverage, while minimizing variations between groups. Additionally, narrower intervals are preferable under the condition that effectiveness is maintained. Furthermore, we introduce PICP Gap, which quantifies the maximum disparity in PICP across different groups, as a fairness evaluation metric.  

\begin{equation}\label{picp-gap}
PICP\ Gap = \max_{s, s' \in [1,S]} |PICP(\mathcal{C}_{\alpha}(X^s)) - PICP(\mathcal{C}_{\alpha}(X^{s'}))|
\end{equation}

\subsection{Results}

In Table \ref{baseline}, we evaluate several commonly used depression recognition algorithms on widely adopted depression datasets. Specifically, for the AVEC 2013 \& 2014 datasets, we focus on the video modality, whereas for DAIC-WOZ and MODMA, we use the audio modality for predictions. Among these datasets, AVEC and DAIC-WOZ contain only gender as a sensitive attribute, while MODMA includes both gender and age. Due to the number of participants in MODMA and the more evident prediction biases observed, we categorize age into two groups: 'Young' representing participants under 30 years old, and 'Old' representing those over 30. Experimental results indicate that biases exist in heterogeneous biases across both age and gender. In the video modality, for example, neither males nor females consistently incur higher (or lower) prediction errors, and the same absence of a clear pattern is observed for speech. Given the limitations of small datasets and the lack of pathological support for depression, we can only conclude that prediction biases are present in depression recognition; we cannot yet determine whether one gender or age group truly has a higher prevalence of depression. Even so, these prediction biases inevitably undermine the fairness of UQ: groups with smaller errors are over-covered, whereas groups with larger errors are under-covered, leading to unequal validity across demographic subgroups.  

Tables \ref{fair} and \ref{fair-age} offer a systematic comparison of several fairness-aware uncertainty-quantification methods across multiple depression datasets, considering two sensitive attributes: gender (present in every dataset) and age (available only in MOMDA). Although all methods meet the global validity requirement at the 90 \% confidence level ($\alpha = 0.1$; PICP $\geq$ 90 \%), marked coverage disparities appear once performance is disaggregated by demographic group. Classical CP and CQR, which were not designed with fairness constraints, consistently under-cover groups with larger prediction errors and over-cover those with smaller errors, producing the widest validity gap. CFQR attempts to mitigate the biases by adding a demographic-parity (DP) adjustment to CQR. DP fairness requires that the post-calibration predictive distributions of different subgroups be similar, but satisfying this constraint widens the prediction intervals, yields only marginal, statistically insignificant gains, and even introduces excess risk that can increase certain gaps. For instance, although CFQR narrows the validity gap of CQR on the MOMDA (age) and DAIC-WOZ datasets, the same adjustment exacerbates disparities on the remaining datasets because the larger intervals overcompensate for random noise rather than systematic bias. In other words, DP fairness definition designed to mitigate prediction bias is ill-suited for uncertainty fairness. BFQR imposes fairness regularization on UQ, but it prioritizes overall coverage so heavily that the resulting intervals become overly conservative; both PICP and MPIW rise, undermining practicality without truly redressing between-group imbalance. This situation is evident in DAIC-WOZ and MOMDA (gender), where BFQR sacrifices tightness while still failing to equalize validity across attributes (as shown in Fig. \ref{fig:triple}).

In contrast, our proposed method strikes a more elegant balance: it maintains PICP at the confident level for every dataset, significantly narrows MPIW, and minimizes the coverage gap between male and female or younger and older subgroups, thereby ensuring EOC fairness while preserving tight intervals. Across different datasets, methods, and sensitive attributes, our approach consistently achieves optimal fairness in prediction. Although most publicly available depression datasets include only gender labels, our approach shows the same validity consistency and minimal coverage disparity for both gender and age on the MOMDA dataset (as shown in Fig. \ref{fig:twosubfigs}), underscoring its robustness to multiple sensitive factors. This robustness is crucial for reducing the risk of unfair outcomes in clinical practice.

\subsection{Parameter Analysis}
\begin{figure}
    \includegraphics[width=0.5\textwidth]{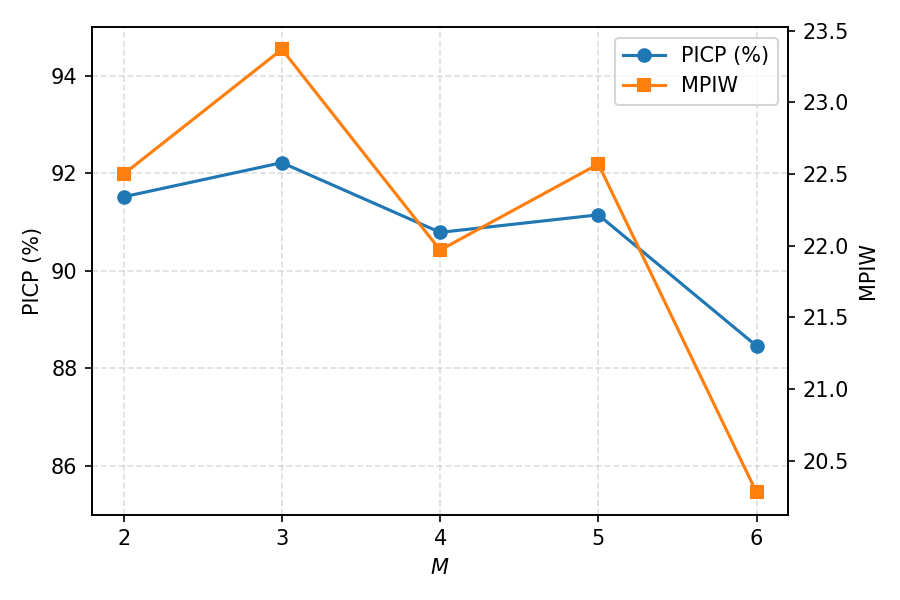}
    \caption{\footnotesize PICP and MPIW with varying $M$ on AVEC 2014.}
    \label{PA:left}
\end{figure}

\begin{figure}
    \includegraphics[width=0.48\textwidth]{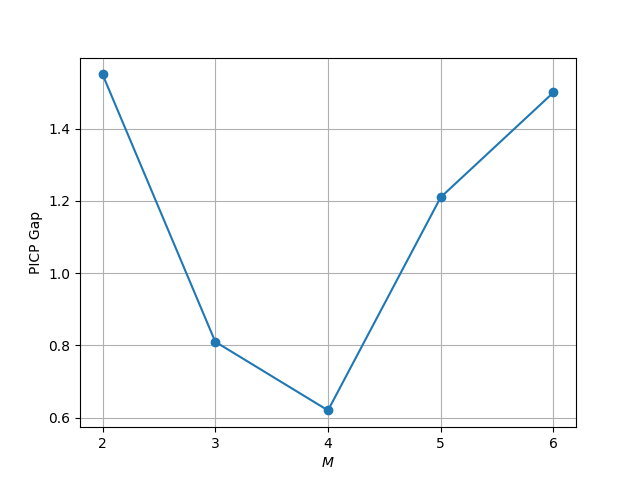}
    \caption{\footnotesize PICP Gap with varying $M$ on AVEC 2014.}
    \label{PA:right}
\end{figure}

$M$ plays a pivotal role in FUQ’s fairness performance. Increasing $M$ decreases the sample sizes of each bin, degrading the reliability of UQ. Decreasing $M$ weakens conditional approximation, drifting toward marginal coverage and producing wider, less informative intervals (under equal confidence, narrower intervals are considered better). In our experiments, we split the calibration set into two disjoint parts to select the optimal hyperparameter $M$. We then fix $M$ and calculate the $\hat{r}_{m,s}$ using the entire calibration set. We do not pursue fairness at all costs; rather, we seek group FUQ while maintaining an overall coverage guarantee. As shown in Figs. \ref{PA:left} and \ref{PA:right}, the optimal value of the parameter ($M = 4$) can be obtained by tuning on the validation set.

\subsection{Limitations and Future Work}
Although FUQ achieves EOC fairness, it still has certain limitations. First, in terms of sensitive attribute grouping, FUQ currently focuses on binary group divisions. If more diverse depression datasets become available in the future, we will further explore the feasibility of multi-group divisions and investigate intergroup communication strategies to achieve an optimized solution. Second, this study does not address fairness in interval prediction for additional modalities or multimodal approaches. Expanding FUQ to incorporate such modalities will be an important direction for future research.

\section{CONCLUSION}
FUQ provides a principled framework for fair and reliable depression prediction within deep learning. Reliability is secured through theoretically guaranteed uncertainty quantification based on group analysis, whereas fairness is attained via a novel group‑based, fairness‑aware optimization. Leveraging the theory of CP, FUQ constructs calibrated prediction intervals with a statistical coverage guarantee. To accommodate group‑specific characteristics, we employ group conditional CP, which adaptively adjusts interval widths across demographic groups within equal-mass binning. On top of these calibrated intervals, we introduce a fairness‑aware optimization strategy for prediction intervals: the objective minimizes the average interval width on the calibration set, subject to an EOC constraint that enforces parity across sensitive‑attribute groups. This formulation delivers uncertainty‑aware fairness while preserving statistical validity, thereby addressing both the ethical and technical demands of depression prediction. Our study constitutes an early step toward fairness‑aware affective computing, and we hope it will inspire future research aimed at ensuring that such technologies are deployed ethically and beneficially in real‑world settings.


\newpage

\bibliographystyle{IEEEtran}
\footnotesize
\bibliography{main}


\end{document}